\documentclass[journal,twoside,web]{ieeecolor}
\usepackage{jsen}
\usepackage[noadjust]{cite}

\usepackage{subcaption}
\usepackage{amsmath,amssymb,amsfonts}
\usepackage{algorithmic}
\usepackage{graphicx}
\usepackage{textcomp}
\usepackage{wrapfig}
\usepackage{multirow}
\usepackage{makecell}
\def\BibTeX{{\rm B\kern-.05em{\sc i\kern-.025em b}\kern-.08em
    T\kern-.1667em\lower.7ex\hbox{E}\kern-.125emX}}
\definecolor{abstractbg}{rgb}{0.89804,0.94510,0.83137}
\begin{document}
\title{DeepMachining: Online Prediction of Machining Errors of Lathe Machines}
\author{Xiang-Li Lu, Hwai-Jung Hsu, Che-Wei Chou, H. T. Kung,  Chen-Hsin Lee, and Sheng-Mao Cheng
\thanks{}}

\IEEEtitleabstractindextext{%
\fcolorbox{abstractbg}{abstractbg}{%
\begin{minipage}{\textwidth}%
\begin{wrapfigure}[11]{r}{3.5in}%
\begin{center}
\includegraphics[width=3.5in, height=1.4in]{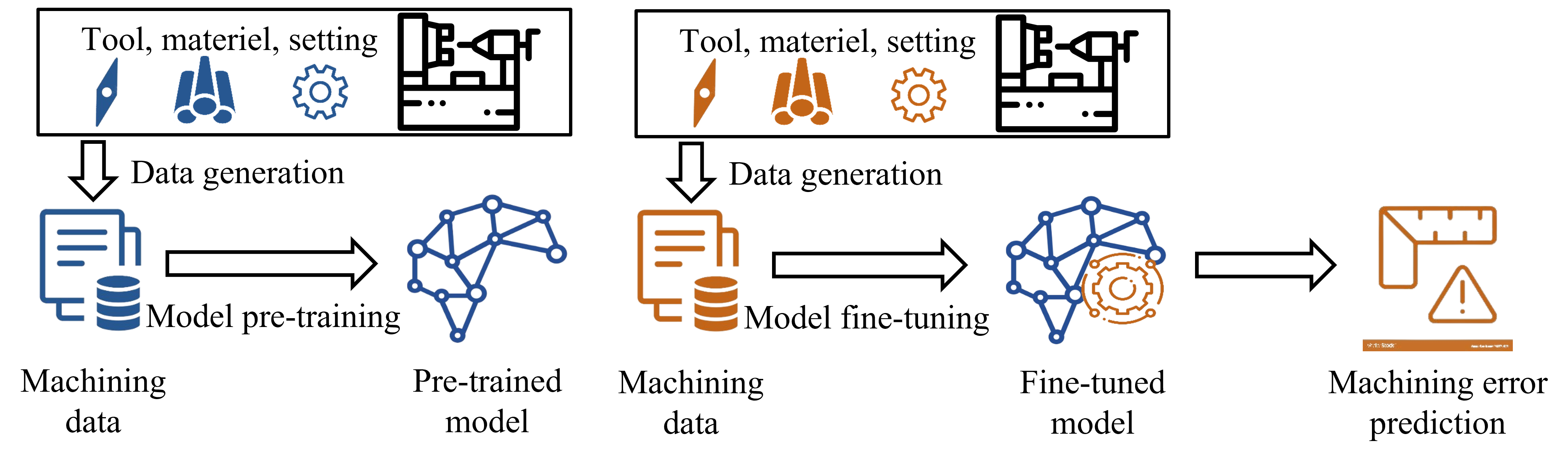}%
\end{center}
\end{wrapfigure}%
\begin{abstract}
We describe DeepMachining, a deep learning-based AI system for online prediction of machining errors of lathe machine operations. We have built and evaluated DeepMachining based on manufacturing data from factories. Specifically, we first pretrain a deep learning model for a given lathe machine's operations to learn the salient features of machining states. Then, we fine-tune the pretrained model to adapt to specific machining tasks. We demonstrate that DeepMachining achieves high prediction accuracy for multiple tasks that involve different workpieces and cutting tools. To the best of our knowledge, this work is one of the first factory experiments using pre-trained deep-learning models to predict machining errors of lathe machines.
\end{abstract}

\end{minipage}}}

\maketitle

\section{Introduction}
\label{sec:introduction}
\IEEEPARstart{T}he structures of modern manufacturing devices are increasingly complex, while tolerance requirements for possible machining errors become more strict. In the manufacturing of high-precision parts, high-quality machining with low errors is essential.

For lathe machines, popular in the manufacturing of precision parts, various machining errors such as geometric tooling, thermal-induced, and load-induced errors \cite{RN941, RN940}, etc., can lead to inaccuracies above the tolerance level of manufactured workpieces, resulting in monetary losses to the manufacturers. Early detection of manufacturing quality degradation and process anomalies \cite{RN879, RN870}, and assessment of the wear of cutting tools in material removal processes \cite{RN908} can help reduce such risks. In particular, implementing real-time monitoring and online machining quality prediction can enhance error detection's efficiency and efficacy.

In recent years, tool condition monitoring (TCM), enabled by sensor technology and artificial intelligence (AI), has been employed to address these needs \cite{RN876}. For example, TCM has been widely used for fault detection and diagnosis (FDD) \cite{RN911, RN888, RN880, RN887}, predictive maintenance (PdM) \cite{RN885, RN884, RN886, RN892}, prognostics and health management (PHM) \cite{RN889, RN890, RN870}, etc. in the manufacturing industry.

Deep-learning-based AI driven by manufacturing data is a promising approach for error detection, given that these data-driven methods have been successful in fields like computer vision and natural language processing \cite{RN886, RN911, RN890, RN887, RN870, RN894}. However, applying deep learning techniques to manufacturing brings new challenges, such as
model generalization for factory environments. For example, real-world machining processes involve a variety of workpiece materials, cutting tools, process recipes, and equipment models. As a result, supervised deep-learning models trained on signals from sensors of specific CNC machines may not apply to other machines. In other words, AI-powered solutions may not generalize to diverse manufacturing environments \cite{RN891}.


We may apply the classical transfer learning approach \cite{marei2021transfer,marei2022cutting,sun2018deep} to address the model generality issue, where a pre-trained models trained on a large labeled dataset is fine-tuned to the target
task. However, acquiring the abnormal data corresponding to machining states that lead to the manufacture of erroneous workpieces is extremely costly in the machinery industry. Thus,
gathering sufficient amounts of high-quality data for training the pretrained model is challenging. 
In addition, the limited computational resources of the CNC machine must be addressed to ensure that the AI solution is deployable. 

\begin{figure}[!t]
    \begin{subfigure}{0.185\textwidth}
        \includegraphics[width=\textwidth, height=1.4in]{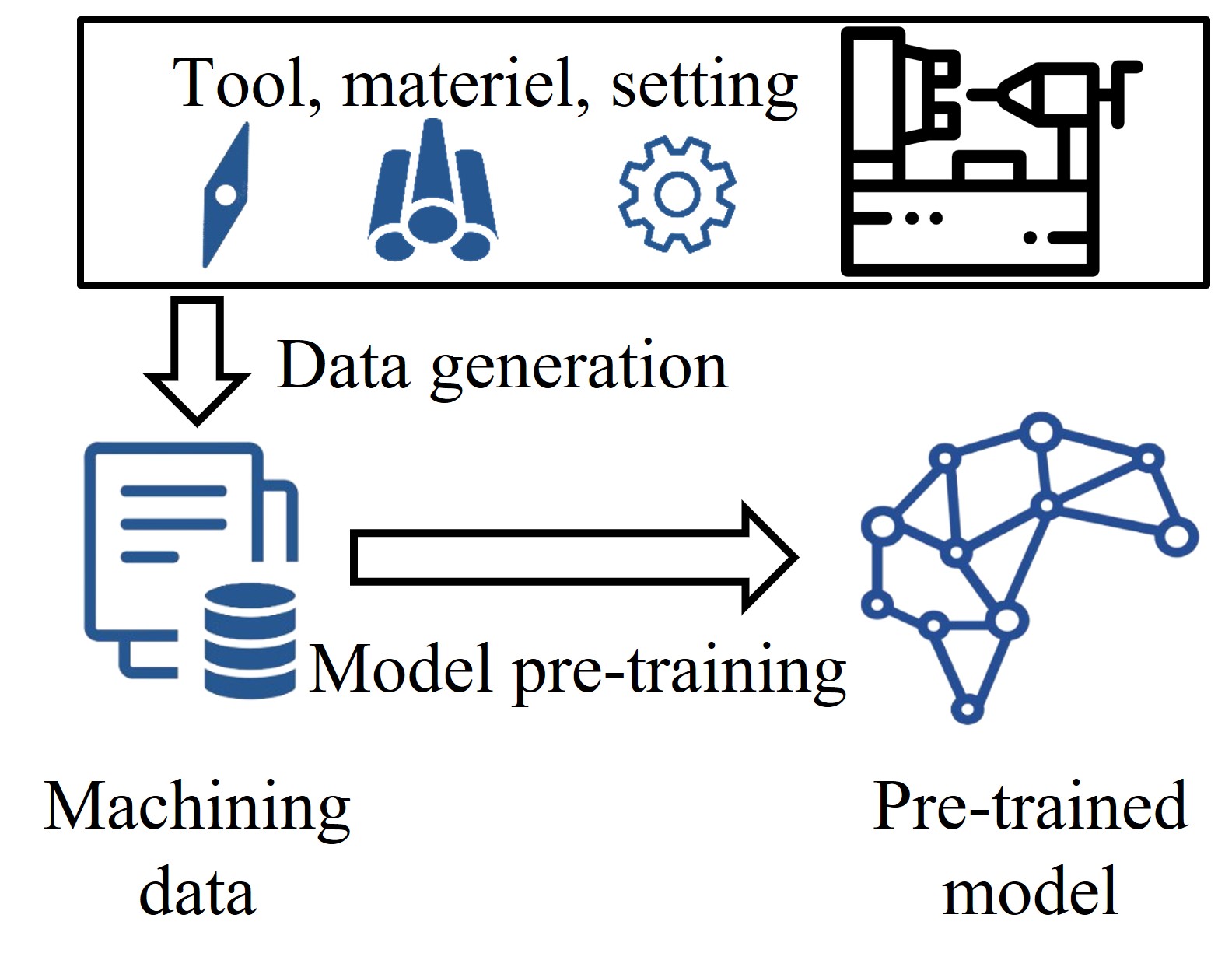}
        \caption{Pre-trained Stage}
        \label{fig:flow_pre-trained}
        \end{subfigure}
    \begin{subfigure}{0.295\textwidth}
        \includegraphics[width=\textwidth, height=1.4in]{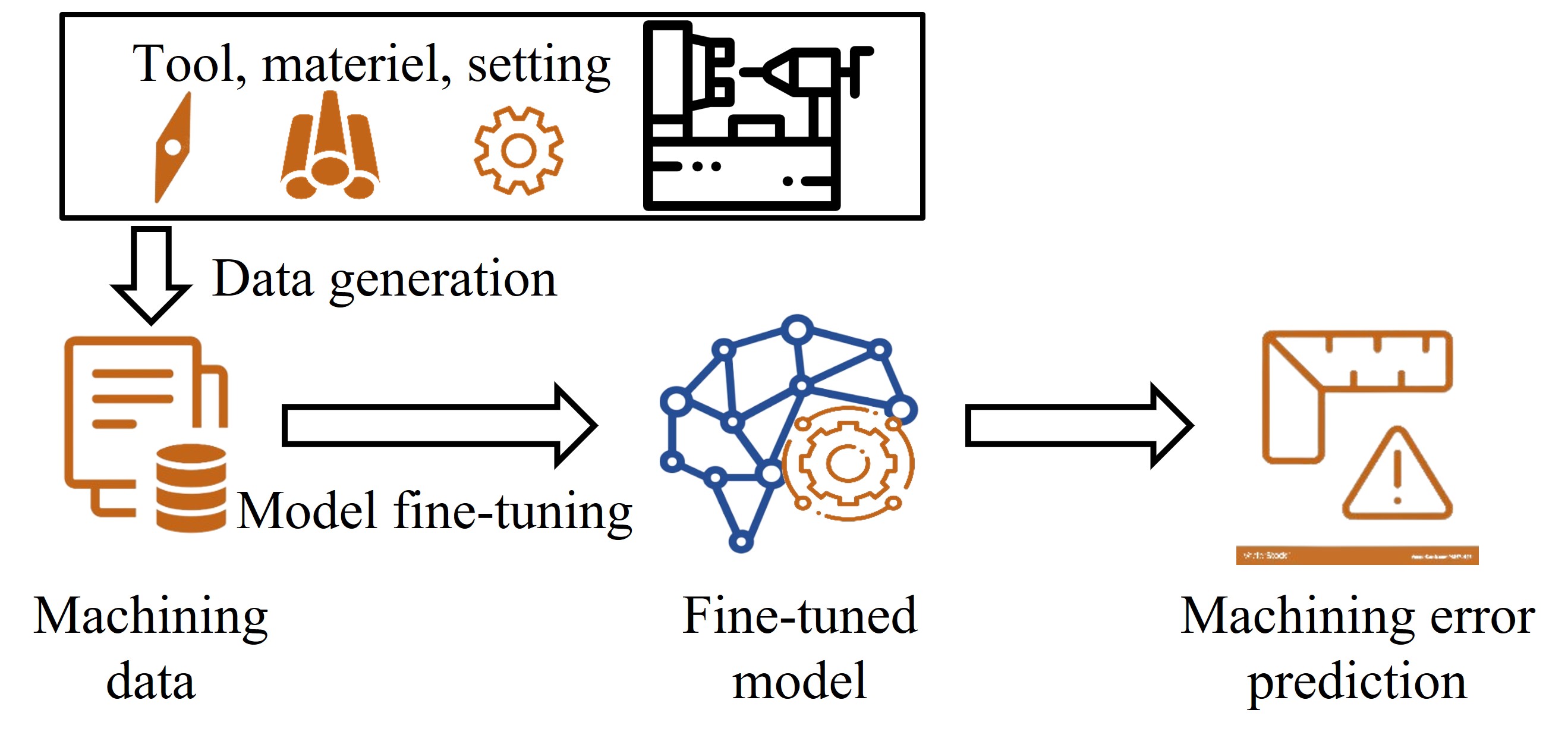}
        \caption{Fine-tuned Stage}
        \label{fig:flow_fine-tuned}
    \end{subfigure}
    \caption{The Process of DeepMachining}
    \label{fig:flow}
\end{figure}

To address these challenges, this paper develops DeepMachining, a deep learning-based AI system, to predict machining errors utilizing the pre-trained model. As Fig.~\ref{fig:flow} shows, the pre-trained model was trained over the lifetime of the cutting tool until it was completely worn out. For model generalization, we perform model pretraining involving multiple spindle speeds. For fine-tuning, we propose a method similar to BitFit \cite{RN863}, which adjusts the model's biases. This allows the pre-trained model to adapt to the target tasks using few-shot learning (typically two-shot). In other words, fine-turning uses data collected from two instances of the target machining task. Merely 6.5\% of the total parameters of the model are fine-tuned in less than 12.5\% of epochs of the model pre-training. Thus, the proposed fine-tuning method not only suits existing machining processes but can also be completed with the limited computational power of the industrial computers in the CNC machines. 

To evaluate the proposed approach in predicting machining errors under various manufacturing settings, we use four machining tasks for the validation. 

The main contributions of this paper are:
\begin{itemize}
    \item The proposed DeepMachining approach, and performed validation showing that, under the approach, we can pre-train a model that can be adapted to various downstream tasks.
    \item A few-shot model fine-tuning method (typically, two-shot) for adaptation to new manufacturing settings.
    \item The useful insight that the fine-tuning required in these manufacturing tasks is basically shifts of model's biases.
    \item An end-to-end factory demonstration of DeepMachining based in real-world factories.
\end{itemize} 

The rest of this paper is organized as follows. Section~\ref{sec:related_work} reviews the related literature on TCM and its applications. Section~\ref{sec:methodology} addresses the DeepMachine framework for online prediction of machining errors. Section~\ref{sec:experiments} details the experiments and analysis using real world machining tasks in factories. Section~\ref{sec:discussion} discusses the limitations and lessons learned in this study. Conclusions are drawn in Section~\ref{sec:conclusion}.

\section{Related Work}
\label{sec:related_work}

\begin{figure*}[!ht]
    \centering
    \begin{subfigure}{0.32\textwidth}
        \includegraphics[width=\textwidth]{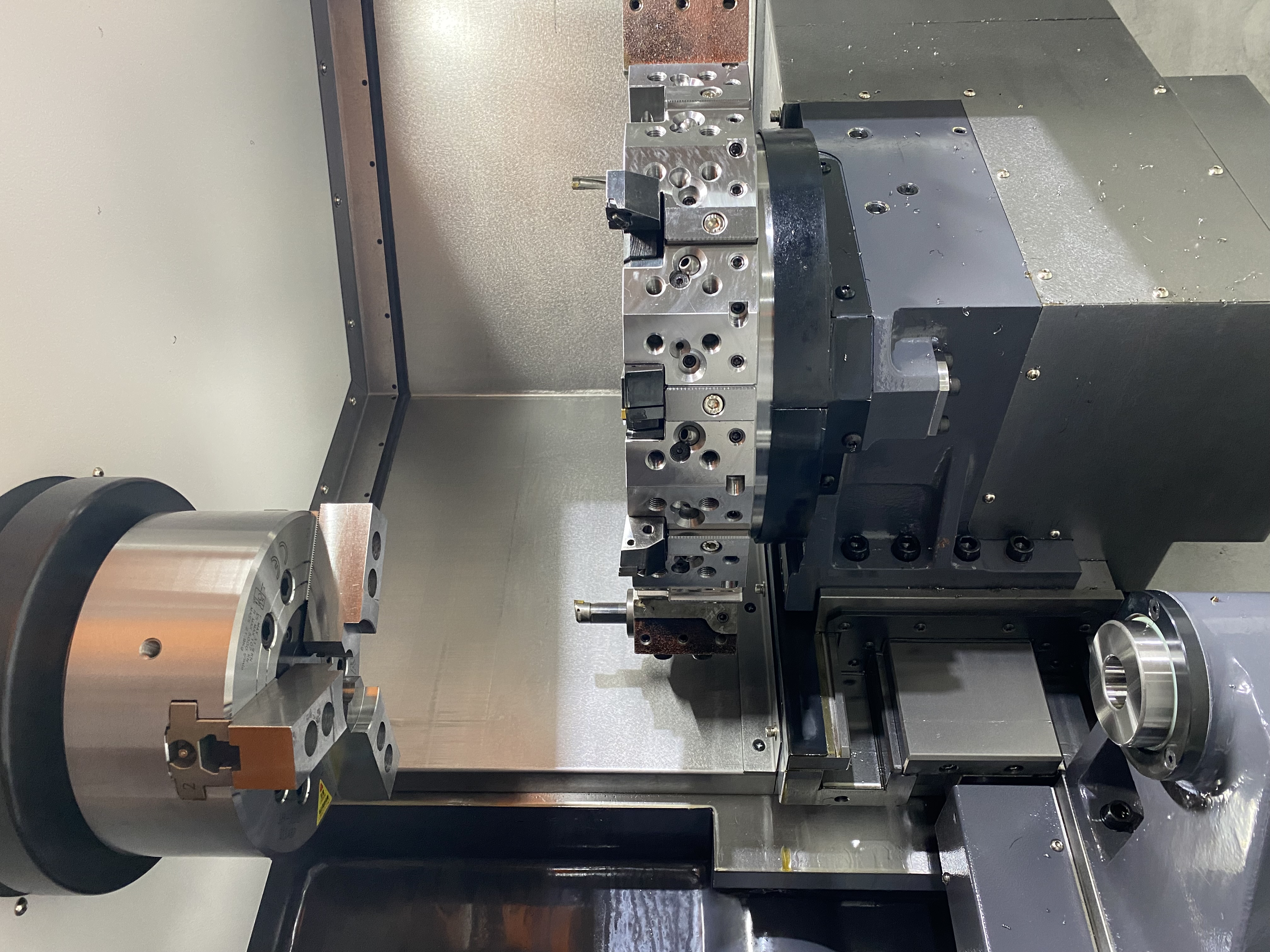}
        \caption{Internal Structure of CNC Lathe Machine}
        \label{fig:enviroment_machine}
    \end{subfigure}
    \begin{subfigure}{0.32\textwidth}
        \includegraphics[width=\textwidth]{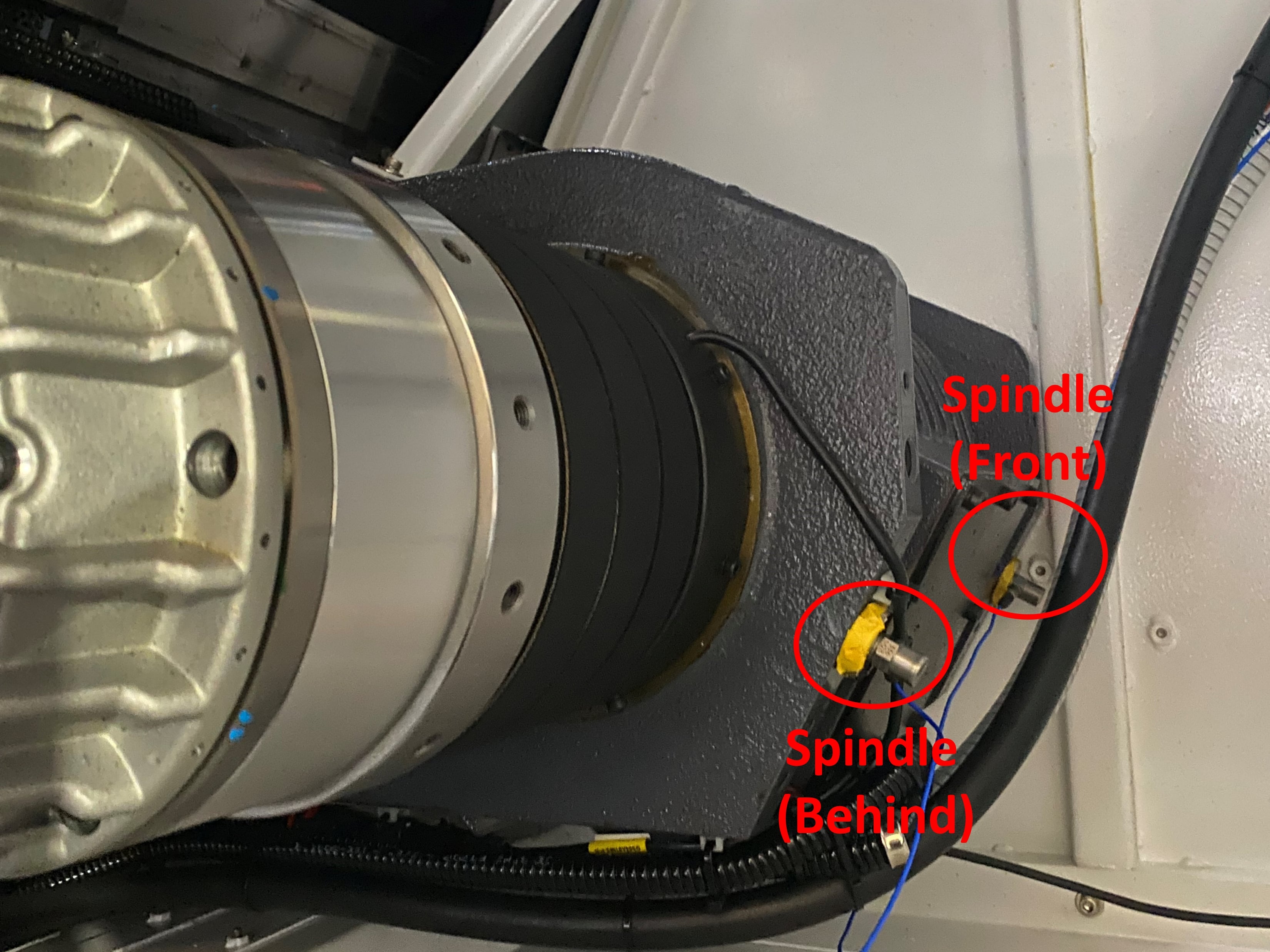}
        \caption{Accelerometer Placement on the Spindle}
        \label{fig:enviroment_spindle}
    \end{subfigure}
    \begin{subfigure}{0.32\textwidth}
        \includegraphics[width=\textwidth]{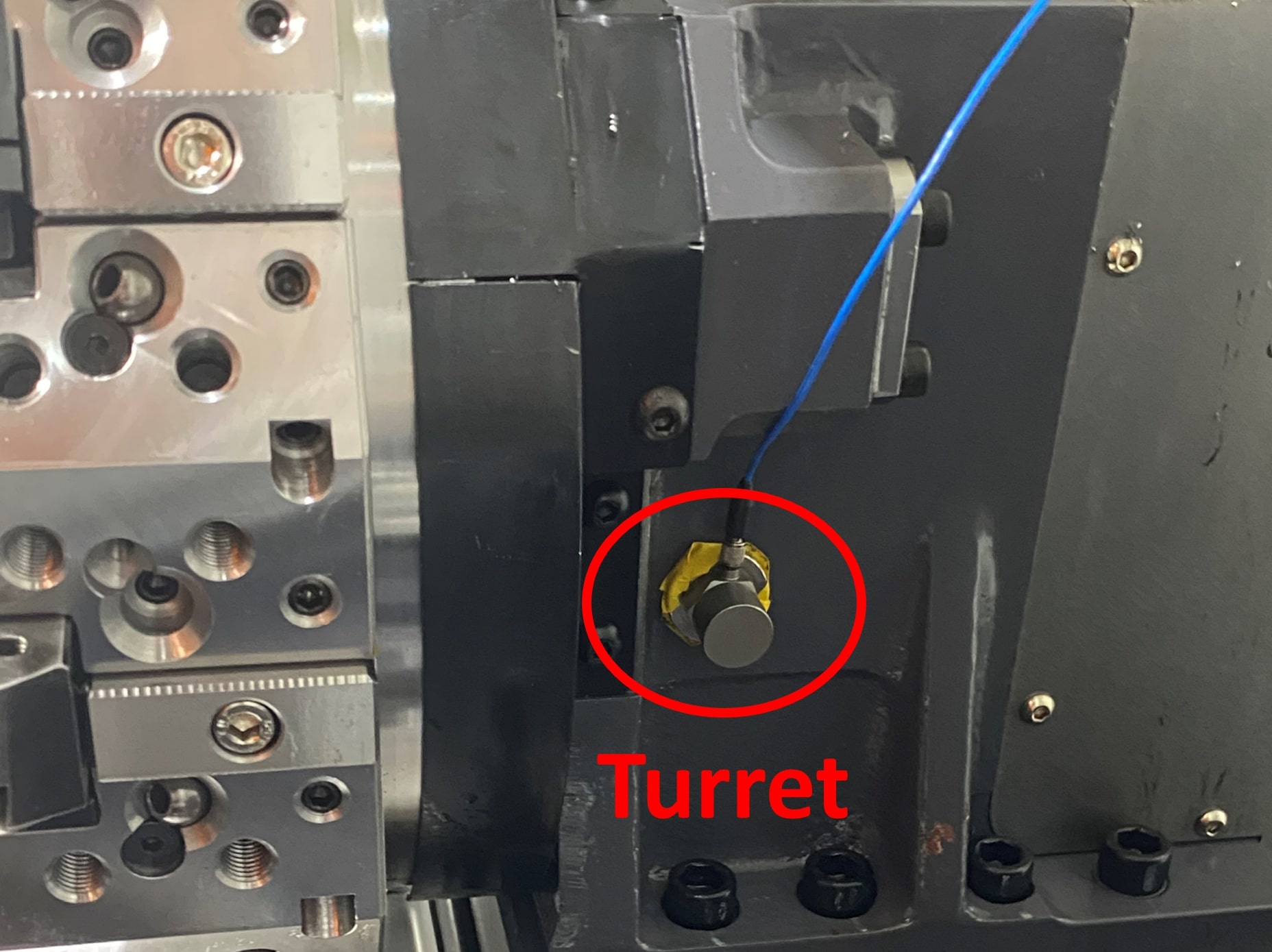}
        \caption{Accelerometer Placement on the Turret}
        \label{fig:enviroment_turret}
    \end{subfigure}
    \caption{The experimental environment of the CNC lathe machine}
    \label{fig:enviroment}
\end{figure*}

Machining error, surface roughness, and tool wear are key quality control metrics in machining processes. Intelligent sensors, including accelerometers, data acquisition encoders, acoustic emission sensors, microphones, dynamometers, and image sensors, are utilized to monitor and diagnose machine health degradation and process anomalies \cite{RN935}. Accelerometer sensors are sensitive and reliable in measuring workpiece dimensions in high precision \cite{RN897,RN898}. Therefore, we adapted accelerometers, DAQ encoder, and microphone to collect manufacturing data building DeepMachining for predicting machining error in this study. 

Traditional machine learning (ML) approaches have been used to predict product quality during CNC machining tasks. Du et al. \cite{RN937} proposed a power spectral density based feature extraction method from spindle vibration and cutting force signals, which accurately predicted product roughness, profile, and roundness using tree-based regressor approaches in hard turning processes. Denkena et al. \cite{RN944} optimized workpiece quality and tool life in cylindrical turning processes by identifying the machined material based on machine learning algorithms. Papananias et al. \cite{RN945} proposed principal component analysis (PCA) based multilayer perceptron (MLP) networks to accurately predict the true position and circularity requirements of a workpiece in an experimental setting. ML approaches could predict well on collected numerous datasets in manufacturing scenarios. However,  ML approaches cannot be used as the kernel for pre-trained models to adapt to various downstream tasks via fine-tuning. 

The advent of deep learning has reformed predictive approaches, enabling end-to-end prediction and diagnosis procedures to enhance CNC machining precision and reliability within smart tool condition monitoring systems  \cite{RN880,RN919}. Huang and Lee \cite{RN921} proposed one-dimensional convolutional neural network (1D-CNN) and sensor fusion approach accurately estimated tool wear and surface roughness for the CNC machining. Hesser and Markert \cite{RN938} demonstrated the feasibility of predicting CNC machine status and tool wear for maintenance plan using artificial neural networks. Proteau et al. \cite{RN916} proposed a variational autoencoder (VAE) regression model to predict the geometrical and dimensional tolerances of workpieces using sensor data in industrial settings. Zhu \cite{zhu2021big} established a long short-term memory (LSTM) model for one-dimensional time series and CNN for two-dimensional images.

Transformer-based networks have been applied to capture association relationships and dependency from vibration signals through the self-attention mechanism for improving performances of the developing models in recent years,  \cite{wu2023transformer,li2022variational,li2022intelligent,bhandari2023implementation,liu2020novel}.  Wu et al. \cite{wu2023transformer} and Li et al. \cite{li2022variational}  studied fault detection and classification in a rotary system with transformer-based models.  Li et al. \cite{li2022intelligent} and Liu et al. \cite{liu2020novel} applied for tool wear prediction in TCM topics. Compared to transformer-based approaches, in this study we utilize  1D-CNN networks with attention mechanism to address the time series data of vibration signals, considering latency and computing power for prompt inference in practice. 

Transfer learning (TL), which learns two types of networks to extract representations, solves cross-domain diagnosis problems with small and imbalanced data \cite{RN880,RN950,RN948,RN951}.  Wang and Gao \cite{RN949} proposed a CNN-based transfer learning technique using vibration analysis for rolling bearing fault diagnosis. Specifically, adapting a pre-trained VGG19 network \cite{RN954}, using non-manufacturing images from ImageNet \cite{5206848} (i.e., model transfer) and transferring the adapted network structure to different fault severity levels and bearing types (i.e., feature transfer). Guo et al. \cite{RN952} proposed a deep convolutional transfer learning network to classify bearing health conditions with unlabeled data. Bahador et al. \cite{RN912} investigated a transfer learning approach for classifying tool wear based on tool vibration in hard turning processes. Ross et al. \cite{RN953} proposed a transfer learning model with Inception-V3 network \cite{szegedy2015going} to detect tool flank wear under distinct cutting environments. 

However, the research gap between practitioners and researchers remains in practice \cite{RN891}. Different processing parameters result in different data distributions, which poses a significant challenge to ML and DL models. Collecting and labeling data with different combinations of materials, tools, process recipes, and machines in practice is difficult and expensive. Furthermore, even if labeled data is obtained from some manufacturing conditions, the resulting predictive models may fail to classify unlabeled data due to intricate manufacturing settings and data distribution discrepancies \cite{RN950,RN953}. Therefore, designing pre-trained models that are easily adaptable to empirical field applications with strong performance is an important topic \cite{RN864,RN865,wolf2020transformers}. Furthermore, fine-tuning on task-specific supervised data enables seamless adaptation to various specific tasks in practical settings \cite{RN900,RN906,RN863,he2021towards}.

\section{Methodology}
\label{sec:methodology}

\begin{figure*}[!ht]
    \centering
    \includegraphics[width=\textwidth]{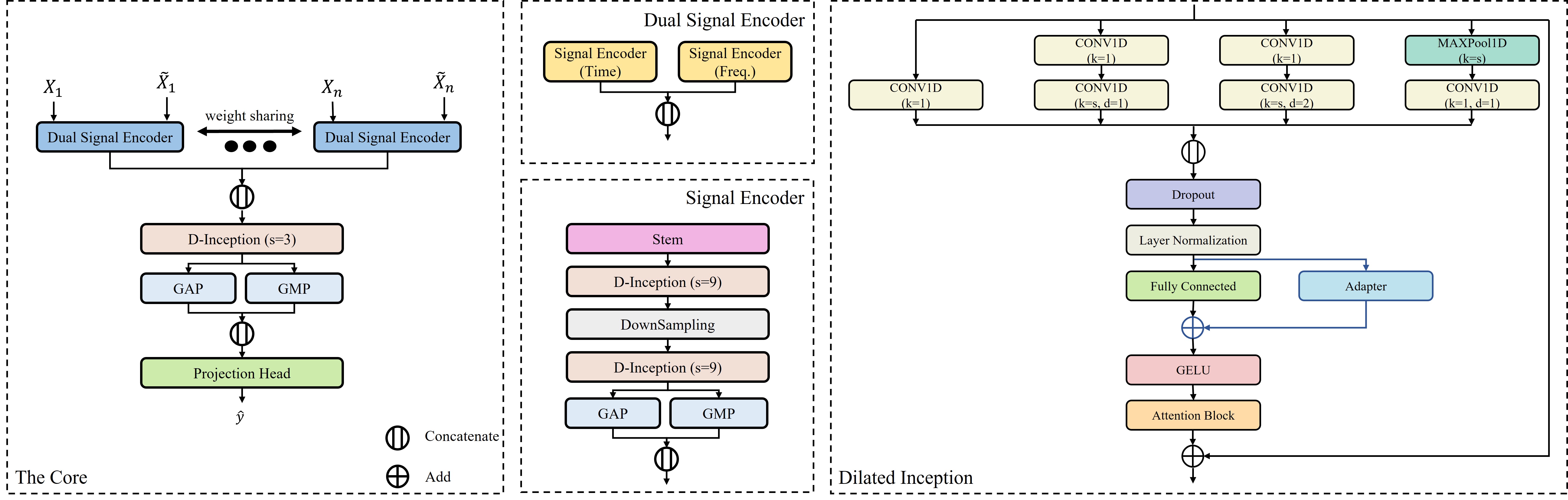}
    \caption{The Structure of The Core of DeepMachining}
    \label{fig:architecture}
\end{figure*}

\subsection{Problem Definition}
The machining error is the difference between the dimension measured after the machining of a workpiece and the target dimension described in the specification. The proposed DeepMachining estimates machining errors under various processing conditions, e.g., different combinations of machining tools and configurations, on CNC lathes without actual measurement. Several factors can impact the machining error of a workpiece. These include the wear condition of cutting tools, the hardness and processing difficulty of the material, the environmental temperature (thermal expansion), and the wear of machine components on the equipment (i.e., the lathe). In order to perceive the factors, accelerometers are installed to collect the vibration signals that occur during the machining process; the machine status, such as the spindle speed and motor current, etc., during the machining process is also recorded. Besides, it's important to note that any specific section of a workpiece can be machined multiple times. In other words, multiple cutting processes may be performed at the same place on a workpiece to achieve the target size. The signals and data generated from multiple machining sessions should be gathered and processed.

\subsection{Machine Settings}
This study conducted experiments on a horizontal CNC lathe machine, which features an internal spindle and three-axis linear guides, as shown in Fig.~\ref{fig:enviroment}. Piezoelectric accelerometers are deployed at three distinct positions, (1) behind the spindle, as depicted in Fig.~\ref{fig:enviroment_spindle}, (2) in front of the spindle, also shown in Fig.~\ref{fig:enviroment_spindle}, and (3) at the base of the tool turret, illustrated in Fig.~\ref{fig:enviroment_turret}, to collect relevant vibration signals. The machine controller records and outputs the spindle speed and current of the drive motors for the spindle and the turret during the machining process, which serves as the machine status.

\subsection{Input Formulation}
In order to predict the machining error $y \in \mathbb{R}^1$, two inputs $\mathcal{X} = \{X_1, X_2, \dots, X_n\} \in \mathbb{R}^{N \times SR \times C_1}$, including the vibration signals and machine status during the machining process, and $\Tilde{\mathcal{X}} = \{\Tilde{X}_1, \Tilde{X}_2, \dots, \Tilde{X}_n\} \in \mathbb{R}^{N \times (\frac{SR}{2}+1) \times C_2}$, the transformation of vibration signals in $\mathcal{X}$ from time domain to frequency domain using Fourier Transform \cite{bracewell1986fourier}, are used. The duration of each input $X_i \in \mathcal{X}$ is one second around the location of the workpiece where the machining error $y$ is measured. $N$ is the number of cuts, $C_1$ and $C_2$ are the number of input channels, and $SR$ indicates the sampling rate of the sensors used in input collection. 


\subsection{The Core of DeepMachining}
The core of DeepMchining, as illustrated in Figure~\ref{fig:architecture}, is a two-stage model handling multiple cuttings across machining processes to estimate the machining error of a workpiece. In the first stage, Dual Signal Encoder, one for $X_n \in \mathcal{X}$ and the other for $\Tilde{X}_n \in \Tilde{\mathcal{X}}$, plays the primary role of feature extraction. Each Signal Encoder contains three blocks, Stem, Downsampling, and Dilated-Inception (D-Inception). Stem retains input representation while reducing its sampling rate. D-Inception and Downsampling are then stacked to extract features. The Signal Encoder simultaneously utilizes Global Average Pooling (GAP) and Global Max Pooling (GMP) to summarize the extracted features and concatenates them to form the output.

In the second stage, hidden states $\mathcal{H} = \{H_1, H_2, \dots, H_n\} \in \mathbb{R}^{N \times 4d}$ extracted from the features produced by the Dual Signal Encoder for each cut are concatenated and processes using an additional D-Inception. $d$ is the dimension of features in the Signal Encoder. Since the Signal Encoder simultaneously concatenates the outputs from the GAP and GMP as its own output. Lastly, the output hidden state from the second stage is used to estimate machining error with a Projection Head, a single-layer feed-forward network.


The details regarding Stem, D-Inception, and Downsampling are explained in the following sections.

\subsubsection{Stem}
$\newline$
\indent Stem reduces input length by lowering the sampling rate using 1D convolution with strides, and thus is learnable. Stem is used at the beginning of the model to reduce the computation costs, with minimal loss of information. The equation of Stem is described as following:
\begin{equation}
    Stem(X_n) = \delta(f^{11}(LN(Dropout(W(X_n)))))
\end{equation}
$LN$ stands for Layer Normalization \cite{ba2016layer}, $f^{11}$ is 1D Convolution whose kernel size equals 11 and strides equals 5, $\delta$ is GELU \cite{hendrycks2016gaussian}, $W \in \mathbb{R}^{C \times d}$ is a learnable weight used to project the input into high dimension space, $C \in \{C_1, C_2\}$ and $d$ are the input and the output channel dimension.

\subsubsection{Dilated Inception}
$\newline$
\indent Dilated Inception (D-Inception) is established for extracting features. The D-Inception incorporates the idea behind Inception, using multiple branches with different convolutions to get features with multiple receptive fields, into transformer-like design, e.g. replacing Batch Normalization \cite{ioffe2015batch} with Layer Normalization \cite{ba2016layer}, replacing activation function ReLU \cite{agarap2018deep} with GELU \cite{hendrycks2016gaussian}, and removing activation functions from the convolution layers in the bottleneck \cite{liu2022convnet}. 

Similar to GoogleLenet \cite{szegedy2015going}, D-Inception utilizes a four-branched Inception, and applies dilated convolution \cite{holschneider1990real} which has the same receptive fields with fewer parameters, instead of normal convolution in some blocks. Applying dilated convolution \cite{holschneider1990real} allows models to be light-weighted with few performance loss. D-Inception is composed of four different one-dimensional (1D) convolution block, normal 1D-convolution with kernel size of 1 (CONV\_1) and $s$ (CONV\_s), and dilated convolution with dilated size of 2 and size of 1 (DCONV\_1) and $s$ (DCONV\_s). The four branches are composed of (1) one CONV\_1, (2) one CONV\_1 and one CONV\_s, (3) one CONV\_1 and one DCONV\_s, and (4) one max-pooling block of size $s$ and one CONV\_1. $s$ is a hyper-parameter indicating the kernel size of some CONV\_s and DCONV\_s in D-Inception. The value of $s$ varies along with the length of the input of a D-Inception block. The outputs of the four branches are concatenated as the features extracted by the convolutional layers. The features are then processed with Dropout \cite{JMLR:v15:srivastava14a} and Layer Normalization \cite{ba2016layer}, and passed to a fully connected layer with activation function of GELU \cite{hendrycks2016gaussian} to form $F \in \mathbb{R}^{L \times d}$ as output. $L$ is the length of the output.

Moreover, similar to CBAM \cite{woo2018cbam}, an Attention Block processing the $F$ added to improve model performance with limited increase of number of model parameters. The Attention Block sequentially calculates a channel attention map $M_c \in \mathbb{R}^{1 \times d}$ and a temporal attention map $M_t \in \mathbb{R}^{L \times 1}$. Our Attention Block is defined as follows:
\begin{equation}
    \begin{split}
        F' &= M_c(F) \otimes F,\\
        F'' &= M_t(F') \otimes F'
    \end{split}
\end{equation}

The channel attention aggregates the temporal information from feature $F$ using average-pooling and max-pooling into $F_{avg}^c$ and $F_{max}^c$. $F_{avg}^c$ and $F_{max}^c$ are then concatenated to feed into a two-layer feed-forward network for calculating attention between features along with channels, and $F'$ is produced as the output of channel attention. The feed-forward network has different activation functions among the layers that GELU ($\delta$) \cite{hendrycks2016gaussian} for the first layer and sigmoid ($\sigma$) for the second. GELU \cite{hendrycks2016gaussian} gives non-linear transformation to the features, and sigmoid scale the output of the attention block between 0 and 1. The calculation of $M_c$ is defined as follows:
\begin{equation}
    \begin{split}
        M_c(F) &= \sigma(MLP(AvgPool(F) || MaxPool(F)))\\
               &= \sigma(W_1(\delta(W_0(F_{avg}^{c} || F_{max}^{c}))))
    \end{split}
\end{equation}
$W_0 \in \mathbb{R}^{\frac{d}{r} \times d}$ and $W_1 \in \mathbb{R}^{d \times \frac{d}{r}}$ are learnable parameters, and $r$ is the reduction ratio used to reduce parameter overhead. 

As with channel attention, temporal attention consumes $F'$ as input, and uses average-pooling and max-pooling on temporal dimension of $F'$ as $F_{avg}^{'t}$ and $F_{max}^{'t}$ separately. $F_{avg}^{'t}$ and $F_{max}^{'t}$ are the concatenated, and processed with 1D convolution with kernel size of 1. The calculation of $M_t$ is defined as follows:
\begin{equation}
    \begin{split}
        M_t(F') &= \sigma(f^{1}(AvgPool(F') || MaxPool(F')))\\
               &= \sigma(f^{1}(F_{avg}^{t} || F_{max}^{t}))
    \end{split}
\end{equation}
$f^{1}$ is 1D Convolution whose kernel size equals 1. $\otimes$ denotes element-wise multiplication.

Attention values are broadcast as follows: channel attention values are broadcast along the temporal dimension, and temporal attention values are broadcast along the channel dimension. Channel and temporal attention weights determine the importance of features and enhance or degrade them accordingly. The output of the attention block is added with the raw input of D-Inception brought by the Residual Connection \cite{he2016deep}. 

\subsubsection{Downsampling}
$\newline$
\indent Similar to Stem, Downsampling reduces input length. However, since Downsampling uses max-pooling to achieve the reduction, it is not learnable. Stacking Downsampling with D-Inception increases the receptive field of D-Inception blocks. The equation of Downsampling is described as follows:
\begin{equation}
    Downsampling(F) = W_1(LN(W_0(MaxPool(F))))
\end{equation}
$LN$ stands for Layer Normalization \cite{ba2016layer}, and $MaxPool$ represents max-pooling. $W_0 \in \mathbb{R}^{4d \times d}$ and $W_1 \in \mathbb{R}^{d \times 4d}$ are learnable weights that shuffle features based on a bottleneck structure.

\subsection{Fine-tuning Method}
\label{ssec:fine-tuning-method}
In order to fine-tune the model for diverse machining tasks, we adopt a methodology inspired by TinyTL \cite{cai2020tinytl}. TinyTL inserts an additional lightweight residual module into the pre-trained model, and trains only the residual module, bias, and regressor during the fine-tuning process.

Instead of the lightweight residual module, we insert an Adapter composed of two feed-forward layers for feature enhancing and shuffling in conjunction with the pre-trained model. Adapters are inserted into D-Inception and Downsampling blocks. During fine-tuning, only the Adapter, Bias, and Projection Head undergo training while all other parameters remain frozen. The Adapter is formally defined as follows:
\begin{equation}
    Adapter(F) = F + W_1(W_0(F))
\end{equation}
$W_0 \in \mathbb{R}^{d \times \frac{d}{r}}$ and $W_1 \in \mathbb{R}^{\frac{d}{r} \times d}$ are learnable weights. $r$ indicates the reduction ratio, and $F \in \mathbb{R}^{L \times d}$ is the feature in D-Inception or Downsampling. 

In practice, workpiece dimensions are measured during CNC machine restarts, process resets, or when operators deem reconfiguration necessary. Then, the operators adjust the cutting tools and machine parameters on the basis of the measurements to ensure the accuracy of subsequent machining. Therefore, the proposed DeepMachining incorporates few-shot learning (typically two-shot) to fine-tune the pre-trained model at these instances to adapt the variations in the machining contexts, parameters, and diverse changes in workpieces and cutting tools. In summary, the core of DeepMachining is a relatively small model with around \textbf{\textit{260,000}} parameters. Only \textbf{\textit{6.5\%}} of the parameters needs to be fine-tuned with \textbf{\textit{12.5\%}} of epochs of the pre-training process for adaptation to various machining tasks and configuration. 

\section{Experiments}
\label{sec:experiments}

\begin{table}
    \caption{Machining Configuration of Datasets}
    \begin{center}
        \begin{tabular}{|c|c|c|c|c|}
            \hline
            \multirow{2}{*}{ Dataset } & \multirow{2}{*}{ Spindle RPM } & Feed Rate & \# of Configuration \\
             & & (mm/rev) & Changes \\ \hline
            WC\_AO-MS & 1100 to 2700 & [0.25,0.1] & 14 \\ \hline
            WC\_TAN-MS & 1600 to 2200 & [0.25,0.12] & 2 \\ \hline
            WC\_TC-AS & 1000 to 2100 & [0.12,0.25] & 3 \\ \hline
        \end{tabular}
    \end{center}
\end{table}

\subsection{Settings}
\textbf{Datasets:} The datasets were collected from three distinct machining tasks, and were named on the basis of the material and coating of the cutting tool, as well as the material of the workpieces under machining. All of the cutting tools used in the experiments were made of Tungsten Carbide (WC). The coatings of the cutting tools include Aluminium Oxide (AO), Titanium Aluminium Nitride (TAN), and Titanium Carbonitride (TC). The materials of the workpieces included Medium-Carbon Steel (MS) and Alloy Steel (AS). On the other hand, except for the vibration and the machine controller signals, adjustments to the machining configurations (e.g. spindle speed, initial tool position) and context changes (e.g. machining dates) along with the machining processes were also recorded. TABLE I shows the summary of each dataset, and the details are described as follows:
\begin{itemize}
    \item WC\_AO-MS: 347 MS workpieces were machined using a tool made of WC and coated with AO. The workpieces were machined on seven different dates, with varying spindle speeds for each date. Besides, according to the judgment of on-site personnel, the cutting tool underwent eight position adjustments to offset its machining precision. Besides the first machining, the acts of machining on the other dates and the too position adjustments are considered as a configuration change. This dataset was used for model pre-training. 
    To evaluate the performance of the pre-trained model, testing dataset was split from the dataset for assessment. The testing dataset was generated in two different ways. First, 80\% of the data was randomly selected for training, and the remaining 20\% for testing. Second, the first 80\% of the dataset (sequenced by machining time) was used for training, and the remaining for testing. In the following sections, the first dataset is named as WC\_AO-MS (Random) and the second one as WC\_AO-MS (Sequential).
    \item WC\_TAN-MS: 87 MS workpieces were machined using a tool made of WC and coated with TAN. The workpieces were machined on two different dates, with varying spindle speeds for each date. Since we plan to fine-tune the pre-trained model to adapt the tool differences in WC\_TAN-MS, each machining date in WC\_TAN-MS is considered as a configuration change.
    To assess whether the pre-trained model can adapt to changes in cutting tools through fine-tuning, few learning is applied for each machining date, i.e. configuration change, as the section~\ref{ssec:fine-tuning-method} describes. In other words, for each date, the first two workpieces are used for model fine-tuning, and the remaining ones are used for testing.
    \item WC\_TC-AS: 34 AS workpieces were machined using a tool made of WC and coated with TC. All the workpieces were machined on the same date. However, there were three instances of configuration change: (1) when the machine started in the morning, (2) when the machine resumed after the lunch break, and (3) when the the cutting tool is adjusted for precision offset. The two workpieces processed after each configuration change were used for model fine-tuning. Subsequent workpieces, processed until the next configuration change or end of the machining, were used for testing. This allowed us to assess whether our fine-tuning approach could adapt to changes in both cutting tools and workpiece materials.
\end{itemize}

\begin{table}
    \caption{Train/Test Split of Pre-trained Datasets}
    \begin{center}
        \begin{tabular}{|c|c|c|c|c|}
            \hline
            \multirow{2}{*}{ Dataset } & Train & Test & Total \\
             & (\#Workpieces) & (\#Workpieces) & (\#Workpieces) \\ \hline
            WC\_AO-MS & \multirow{2}{*}{ 277[$277$] } & \multirow{2}{*}{ 70[70] } & \multirow{2}{*}{ 347 } \\
            (Random) & & & \\ \hline
            WC\_AO-MS & \multirow{2}{*}{ 281[$277$,2,2] } & \multirow{2}{*}{ 66[11,7,48] } & \multirow{2}{*}{ 347 } \\
            (Sequential) & & & \\ \Xhline{1pt}
        \end{tabular}
    \end{center}
    \label{table:Datasets1}
\end{table}

\textbf{Evaluation Metrics:} The performance of our method is evaluated by Root Mean Square Error (RMSE), Mean Absolute Error (MAE) and Pearson Correlation (CORR). MAE and RMSE are both used to assess whether a model accurately estimates actual machining errors. RMSE is more sensitive to outliers compared to MAE, and MAE is considered more intuitive to the domain experts. 

CORR is used to observe whether the model's estimation of machining error is correlated with the actual ones. Since in certain machining process demanding high precision, the variations of machining errors are small. A model can give machining error estimation in a small value interval to get small MAE and RMSE. However, in such circumstances, if the model is not trully capable of predicting the machining error, the CORR would be low. In other words, CORR assists us in distinguishing whether a model really learns the relationships between the signals during machining process and the machining errors. On the other hand, low CORR with high MAE or RMSE indicates that the estimation made by the model is biased.

\textbf{Baselines}: Three methods were chosen as the baseline methods for comparison. 
\begin{itemize}
    \item SVR: Support vector regression (SVR) is a kernel-based machine learning model for regression tasks \cite{cortes1995support}. SVR utilizes kernel functions to identify key data points influencing the regression hyperplane and achieve efficient high-dimensional space mapping. In CNC machine applications, SVR is applied for engineering optimization problems such as optimizing surface roughness and cutting forces in milling \cite{yeganefar2019use}, and controlling the motor current of machine tool drives \cite{schwenzer2020support}, etc. In this study, referring to the \cite{sayyad2022tool}, the statistical features of vibration, spindle speed, and motor current signals were processed as input of SVR for model training and machining error inferences.
    \item 1D-CNN: The one-dimensional convolutional neural network (1D-CNN) is commonly employed for the analysis of time series data. In this study, we adopted the 1D-CNN method proposed by Huang and Lee \cite{RN921} as a representative for the baseline comparison. In Huang and Lee's approach \cite{RN921}, 1D-CNN was adopted in conjunction with a sensor fusion technique to accurately estimate tool wear and surface roughness in CNC machining. In this study, the vibration signals were utilized as inputs to the model for the estimation of machining errors. 
    \item 2D-CNN: Once a series of vibration or sound signals is transformed into spectrograms, a visual representation of the spectrum of frequencies of a signal as it varies with time, they can be analyzed using a two-dimensional convolutional neural network (2D-CNN). In this study, the approach delivered by Liao et al. \cite{liao2021manufacturing} was introduced as a representative of 2D-CNN approaches for the baseline comparison. Liao et al. processed sound signals using Short-Time Fourier Transform (STFT) \cite{bracewell1986fourier}, and transformed the spectrum variations over time into spectrograms to predict specific machining configurations \cite{liao2021manufacturing}. Liao et al. fine-tuned a VGG16-based model, which was pre-trained on ImageNet \cite{5206848} to accept spectrograms as input. In this study, we adopted the model framework proposed by Liao et al. \cite{liao2021manufacturing}. However, instead of sound signals, we transformed the vibration signals into spectrograms using STFT as the model inputs. We then fine-tuned the model accordingly for the estimation of machining errors.
\end{itemize}

\begin{table}
    \caption{Train/Test Split of Adapted Datasets}
    \begin{center}
        \begin{tabular}{|c|c|c|c|c|}
            \hline
            \multirow{2}{*}{ Dataset } & Train & Test & Total \\
             & (\#Workpieces) & (\#Workpieces) & (\#Workpieces) \\ \hline
            WC\_TAN-MS & 4[2,2] & 83[37,46] & 87 \\ \hline
            WC\_TC-AS & 6[2,2,2] & 28[5,2,19] & 34 \\ \hline
        \end{tabular}
    \end{center}
    \label{table:Datasets2}
\end{table}

\textbf{Devices}: We pre-trained the core of DeepMachining on a workstation equipped with an AMD Ryzen Threadripper processor 3990X (256M Cache, 2.9 GHz), 256 GB RAM, and an NVIDIA Quadro RTX 8000 (48 GB DDR6 RAM) using TensorFlow. AdamW was employed as the optimizer, with a learning rate of 0.001, a batch size of 512, and 512 epochs. For 2-shot tuning, the core was executed on a host featuring an Intel Xeon Silver 4210 processor (13.75M Cache, 2.2 GHz), 256 GB RAM, and an NVIDIA RTX 2080 Ti (11 GB DDR6 RAM) using TensorFlow. AdamW was again employed as the optimizer, with a learning rate of 0.00001, a batch size of 32, and 64 epochs. The 2-shot tuning time on GPU was 2.5 minutes, with an inference time of 0.026 seconds. The 2-shot tuning time on CPU was 35 minutes, and the inference time was 0.036 seconds.

\begin{figure*}[!ht]
    \begin{subfigure}{1\textwidth}
        \includegraphics[width=\textwidth]{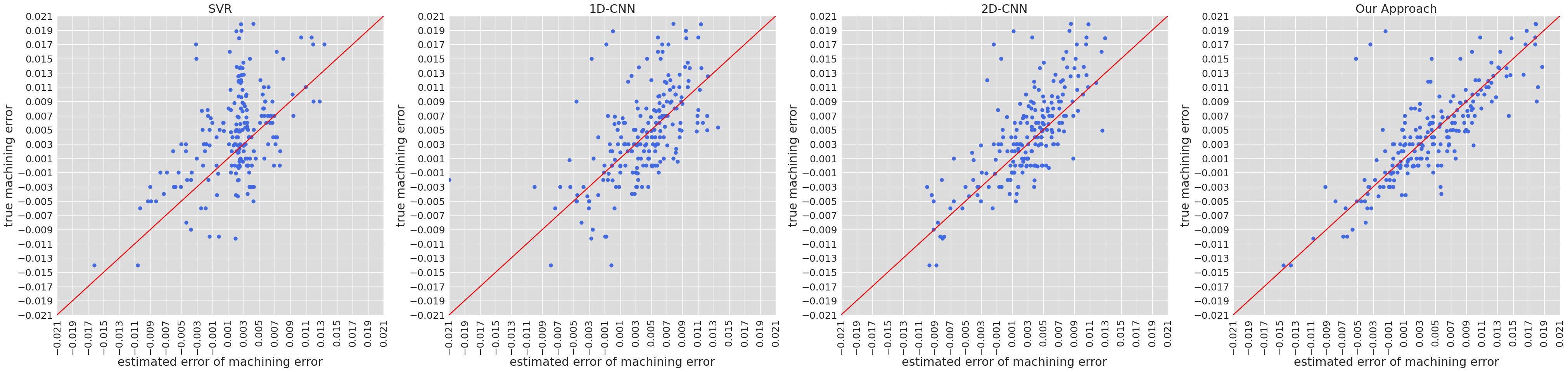}
        \caption{WC\_AO-MS (Random)}
        \label{fig:scatter_plot_pre_a}
        \end{subfigure}
    \begin{subfigure}{1\textwidth}
        \includegraphics[width=\textwidth]{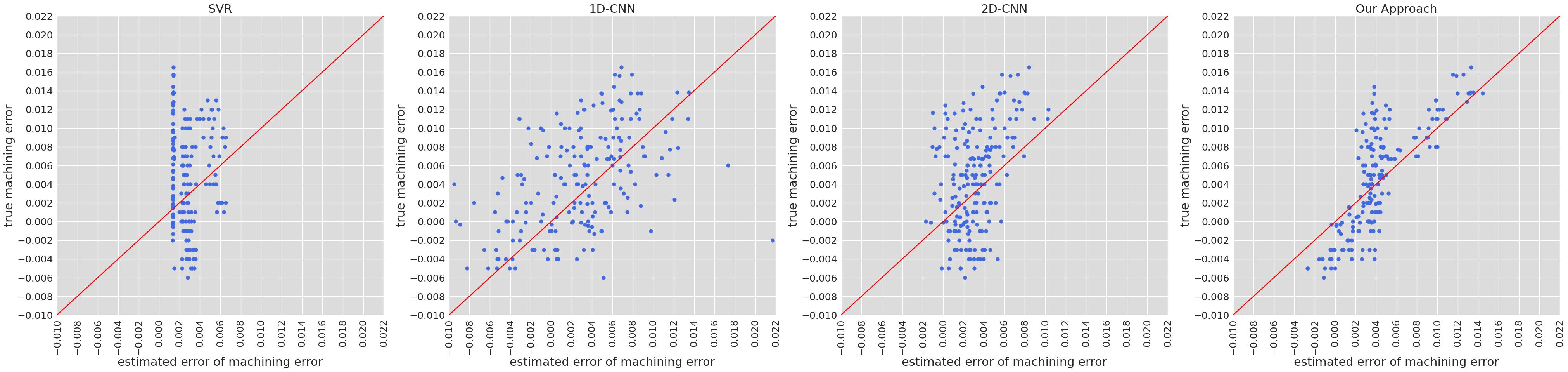}
        \caption{WC\_AO-MS (Sequential)}
        \label{fig:scatter_plot_pre_b}
    \end{subfigure}
    \caption{Scatter plots for the actual machining errors vs. the ones estimated by different methods on our Pre-trained Dataset.}
    \label{fig:scatter_plot_pre}
\end{figure*}

\begin{table}
    \caption{Performance Comparison of Pre-trained Dataset}
    \begin{center}
        \begin{tabular}{|c|c|ccc|}
            \hline
            Dataset & Method & MAE & RMSE & CORR \\ \hline
            \multirow{4}{*}{ WC\_AO-MS } & SVR & 0.0049 & 0.0062 & 0.5052 \\
            & 1D-CNN & 0.0039 & 0.0053 & 0.5864 \\
            \multirow{2}{*}{ (Random) } & 2D-CNN & 0.0036 & 0.0049 & 0.7353 \\
            & \textbf{Our Approach} & \textbf{0.0026} & \textbf{0.0040}  & \textbf{0.8020} \\ \hline
            \multirow{4}{*}{ WC\_AO-MS } & SVR & 0.0050 & 0.0061 & -0.0463 \\
            & 1D-CNN & 0.0045 & 0.0057 & 0.4722 \\
            \multirow{2}{*}{ (Sequential) } & 2D-CNN & 0.0043 & 0.0052 & 0.4029 \\
            & \textbf{Our Approach} & \textbf{0.0028} & \textbf{0.0036} & \textbf{0.7754} \\ \hline
        \end{tabular}
    \end{center}
    \label{table:Results1}
\end{table}

\subsection{Evaluation and Comparison}
Initially, we validated the model's performance on the pre-trained dataset WC\_AO-MS as shown in Table~\ref{table:Datasets1}. As demonstrated in Table~\ref{table:Results1}, whether the testing set is generated randomly WC\_AO-MS (Random), or sequentially WC\_AO-MS (Sequential), our approach surpasses all the baseline methods across various metrics. SVR presents the weakest performance among all the methods, as highlighted by the highest MAE and RMSE, coupled with the lowest CORR Notably, The CORR of SVR is close to 0.5 in WC\_AO-MS (Random) but declines significantly to nearly 0 in WC\_AO-MS (Sequential). This indicates its limited robustness as machining progresses. 2D-CNN outperforms 1D-CNN's in most metrics, and exhibits inferior in CORR for WC\_AO-MS (Sequential). Besides, in comparison to WC\_AO-MS (Random), both 1D-CNN and 2D-CNN demonstrate a substantial increase in estimation errors and decrease in CORR on WC\_AO-MS (Sequential). 

In reality, only first few workpieces processed can be used for model fine-tuning. An approach that cannot perform well with sequential workpieces in production is not practical. In comparison of WC\_AO-MS (Random), our approach shows only a slight uptick in estimation errors and a limited reduction in CORR in WC\_AO-MS (Sequential). Our results demonstrate that our approach can sustain its predictive performance during continuous machining. 

Fig.~\ref{fig:scatter_plot_pre} shows the relationships between the actual machining errors of the testing set (the y-axis) and the estimation of machining errors made by each method (the x-axis). Fig.~\ref{fig:scatter_plot_pre_a} is for WC\_AO-MS (Random), and Fig.~\ref{fig:scatter_plot_pre_b} is for WC\_AO-MS (Sequential). The scales for the plots on the same dataset are equivalent, and the red line in the middle is the identical line representing a perfect match between the actual values and the corresponding estimations.

Fig.~\ref{fig:scatter_plot_pre_b} indicates that SVR only estimates machining errors within a limited range of 0.003 to 0.007 and shows no correlation between the estimated values and actual ones on WC\_AO-MS (Sequential). Both 1D-CNN and 2D-CNN provide estimates with medium correlation to the actual values in both datasets. From Fig.~\ref{fig:scatter_plot_pre_a}, we can see that the estimates made by 1D-CNN have higher variance, as evidenced by the dots being more widely distributed around the identical line. However, in Fig.~\ref{fig:scatter_plot_pre_a}, while 1D-CNN maintains the same tendency in estimation, the estimates made by 2D-CNN seem to be limited by an invisible lower bound around -0.001 mm (the lowest value of actual machining errors in WC\_AO-MS (Sequential) is -0.006 mm). As illustrated in Fig.~\ref{fig:scatter_plot_pre}, compared to 1D-CNN and 2D-CNN, our approach exhibits smaller estimation errors and better correlation for both the WC\_AO-MS (Random) and  WC\_AO-MS (Sequential) datasets. Additionally, the dots in the plots follow the identical line in a more compact manner. 

\begin{figure*}[!ht]
    \begin{subfigure}{1\textwidth}
        \includegraphics[width=\textwidth]{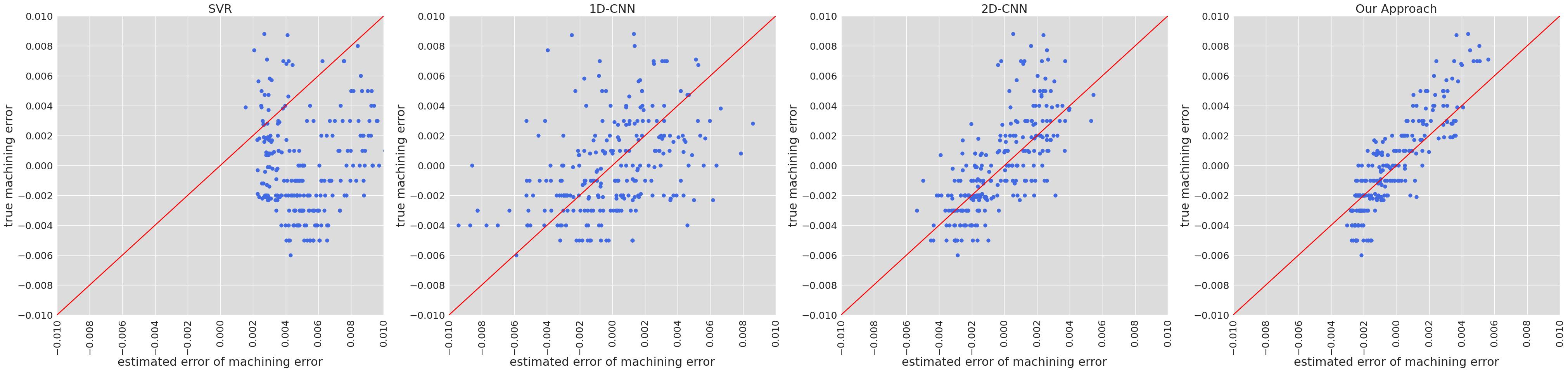}
        \caption{WC\_TAN-MS}
        \label{fig:scatter_plot_adapted_a}
    \end{subfigure}
    \begin{subfigure}{1\textwidth}
        \includegraphics[width=\textwidth]{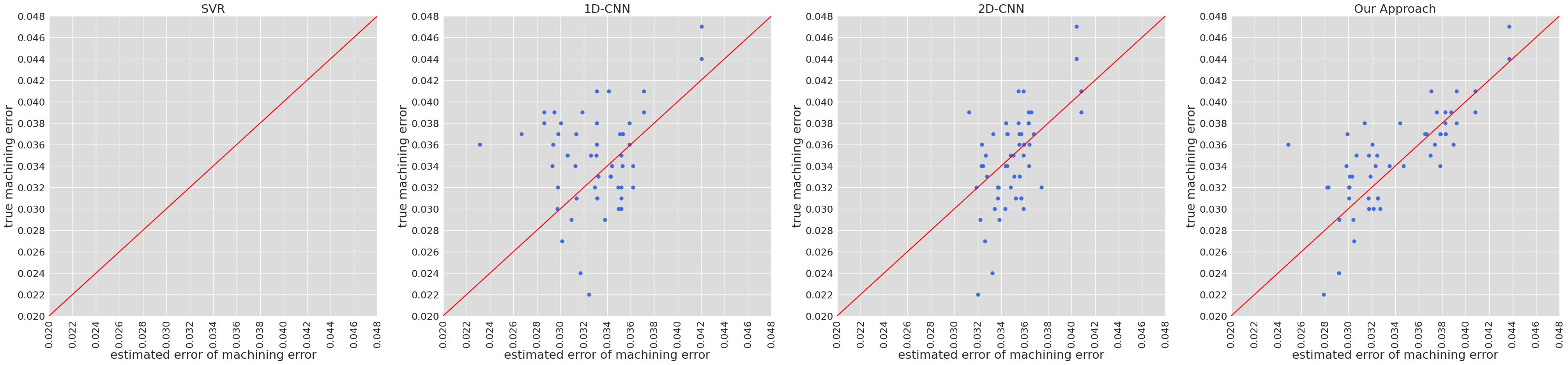}
        \caption{WC\_TC-AN}
        \label{fig:scatter_plot_adapted_b}
    \end{subfigure}
    \caption{Scatter plots for the actual machining errors vs. the ones estimated by different methods on our adapted Dataset.}
    \label{fig:scatter_plot_adapted}
\end{figure*}

\begin{table}
    \caption{Performance Comparison of Adapted Dataset}
    \begin{center}
        \begin{tabular}{|c|c|ccc|}
            \hline
            Dataset & Method & MAE & RMSE & CORR \\ \hline
            \multirow{4}{*}{ WC\_TAN-MS } & SVR & 0.0056 & 0.0064 & 0.0319 \\
            & 1D-CNN & 0.0027 & 0.0034 & 0.4058 \\
            & 2D-CNN & 0.0016 & 0.0022 & 0.7240 \\
            & \textbf{Our Approach} & \textbf{0.0013} & \textbf{0.0016} & \textbf{0.8838} \\ \hline
            \multirow{4}{*}{ WC\_TC-AN } & SVR & 0.2709 & 0.3247 & 0.0339 \\
            & 1D-CNN & 0.0041 & 0.0052 & 0.2518 \\
            & 2D-CNN & 0.0029 & 0.0037 & 0.6010 \\
            & \textbf{Our Approach} & \textbf{0.0024} & \textbf{0.0032} & \textbf{0.7599} \\ \hline
        \end{tabular}
    \end{center}
    \label{table:Results2}
\end{table}

As Table~\ref{table:Datasets2} shows, WC\_TAN-MS is used to show the adaptability for the models in machining with identical material in workpieces but different in tools, and WC\_TC-AN is used to demonstrate the adaptability of the models involving different materials in both workpieces and tools. As shown in Table~\ref{table:Results2}, The CORR of SVR is close to zero, which reveals that SVR fails to adapt the differences in machining condition (no matter in workpieces or tools) for machining error prediction. On the other hand, 2D-CNN outperforms 1D-CNN in both the datasets. However, it's important to note that our approach surpasses all the other methods, emerging as the best across all the metrics.

From Fig.~\ref{fig:scatter_plot_adapted}, SVR's estimation shows no correlation to the actual data, and even out of the plotting range with high errors on WC\_TC-AN. For the adapted dataset, 1D-CNN and 2D-CNN still perform better than SVR but worse than our approach, and both of them meet an invisible lower bound around 0.03 in estimation on WC\_TC-AN. As a result, our approach demonstrates great adaptability and generality while being applied to the datasets generated with different materials in workpieces or tools.

\section{Discussion}
\label{sec:discussion}
We have conducted tests on DeepMachining across different types of products in different manufacturing factories. Some lessons learned are addressed below. (1) Sensor installation position: the placement of sensors on CNC machines is crucial. Incorrect positions may lead to ineffective signal reception, resulting in weak signal amplitudes or noise caused by sensor wire pulling. (2) Sensor sampling rate: different sampling rates between pre-trained and fine-tuned stages can impact the model's performance. To ensure the accurate functionality of DeepMachining, the sampling rate shall be identical at the pre-trained and the fine-tuned stage. (3) Decimal precision: for workpieces with high machining precision, low measurement precision, e.g., measuring a workpiece with a required tolerence of 0.001mm but with a precision of only 0.01mm, it cannot accurately reflect the differences in machining error among workpieces. This lack of precision in data can hinder the model's ability to give precise estimation of machining errors.

In this study, we have only tested the performance of DeepMachining in outer diameter machining tasks on a lathe machine, limiting its application scope to such tasks. However, CNC machining typically encompasses a range of processes, including internal cutting and drilling. Additionally, CNC machines come in various types, such as milling and planing machines. More data from different machining tasks need to be collected and investigated to understand how DeepMachining can be applied to a wider range of machining tasks. In addition to the machining error metric, the workpiece's roughness is another important metric commonly used in CNC machining. However, DeepMachining cannot tackle roughness currently. Addressing this issue by DeepMachining is necessary in the future. 
In addition to the machining error metric, another important metric commonly used in CNC machining is the workpiece's roughness. Currently, DeepMachining has not yet been applied to handle this aspect. However, it's necessary for DeepMachining to address this issue in the future. 
Furthermore, applying DeepMachining in learning CNC machining representations for predicting the remaining useful life of cutting tools is also a significant practical research topic. Ultimately, the research goal is to build the entire DeepMachining system as an open-source intelligent manufacturing platform to have a greater impact on the CNC machinery industry.

\section{Conclusion}
\label{sec:conclusion}
This paper proposes DeepMachining, a deep learning-based approach for estimating machining errors for outer diameter processing in horizontal CNC lathe machines. DeepMachining consists of two stages: 1) pre-training a deep learning model and 2) employing few-shot learning (typically two-shot learning) to adapt the pre-trained model to new machining tasks for estimating machining errors. 

The core of DeepMachining is pre-trained with machining data from one single tool and workpiece material where the data span from a new tool to completely worn-out, i.e., covering the whole life cycle of a tool. The pre-trained model can be adapted to machining tasks with different tools and workpiece materials by applying the collected data from the first two workpieces in new machining tasks for fine-tuning, also known as two-shot learning, whenever the machining configuration is changed. In summary, based on practical experiments, DeepMachining surpasses all the other baseline methods in terms of estimation accuracy and generality. 

Besides, the core of DeepMachining is relatively small with around \textit{260,000} parameters, and only \textit{6.5\%} of the parameters need to be fine-tuned with \textit{12.5\%} of epochs compared to pre-training for adaptation to machining tasks with different tools, workpieces, and machining configurations. DeepMachining can operate under limited computational resources, while also aligning with the processes of the manufacturing industry using CNC machinery. It is believed that DeepMachining will be a paradigm shift for CNC manufacturers and customers, guided by the principles of deep learning in artificial intelligence and intelligent manufacturing. As manufacturing processes become more complex, more potential relationships exist within manufacturing data among processes, workpieces, and machines. Vast amounts of textual data from machine logs correspond to the quality and yield of workpieces. A large language model (LLM) can excel in a range of natural language processing tasks that understand the machine logs and process recipes. Utilizing LLM in complicated production contexts can help manufacturers suffer less machine downtime and better product quality. Therefore, studying LLM in intelligent manufacturing is a crucial area of future study.



\section*{Acknowledgment}
The authors would like to thank Victor Taichung Machinery Works Co. and their employees for the domain knowledge and the data shared in this project. We would also like to acknowledge the financial support of the Taiwan AI Academy.

\bibliographystyle{IEEEtran}
\bibliography{main}

\begin{IEEEbiography}[{\includegraphics[width=1in,height=1.25in,clip,keepaspectratio]{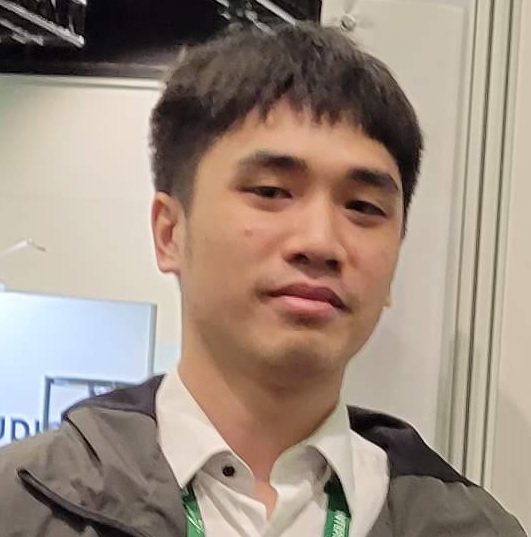}}]{Xiang-Li Lu} received his B.E. and M.S. degrees in Information Engineering and Computer Science from Feng Chia University, Taiwan, in 2020 and 2024. He is currently serving as a research assistant of professor Hwai-Jung Hsu in the Department of Information Engineering and Computer Science at Feng Chia University. His research interests include graph neural networks, speech analysis, computer vision, and deep learning.
\end{IEEEbiography}

\begin{IEEEbiography}[{\includegraphics[width=1in,height=1.25in,clip,keepaspectratio]{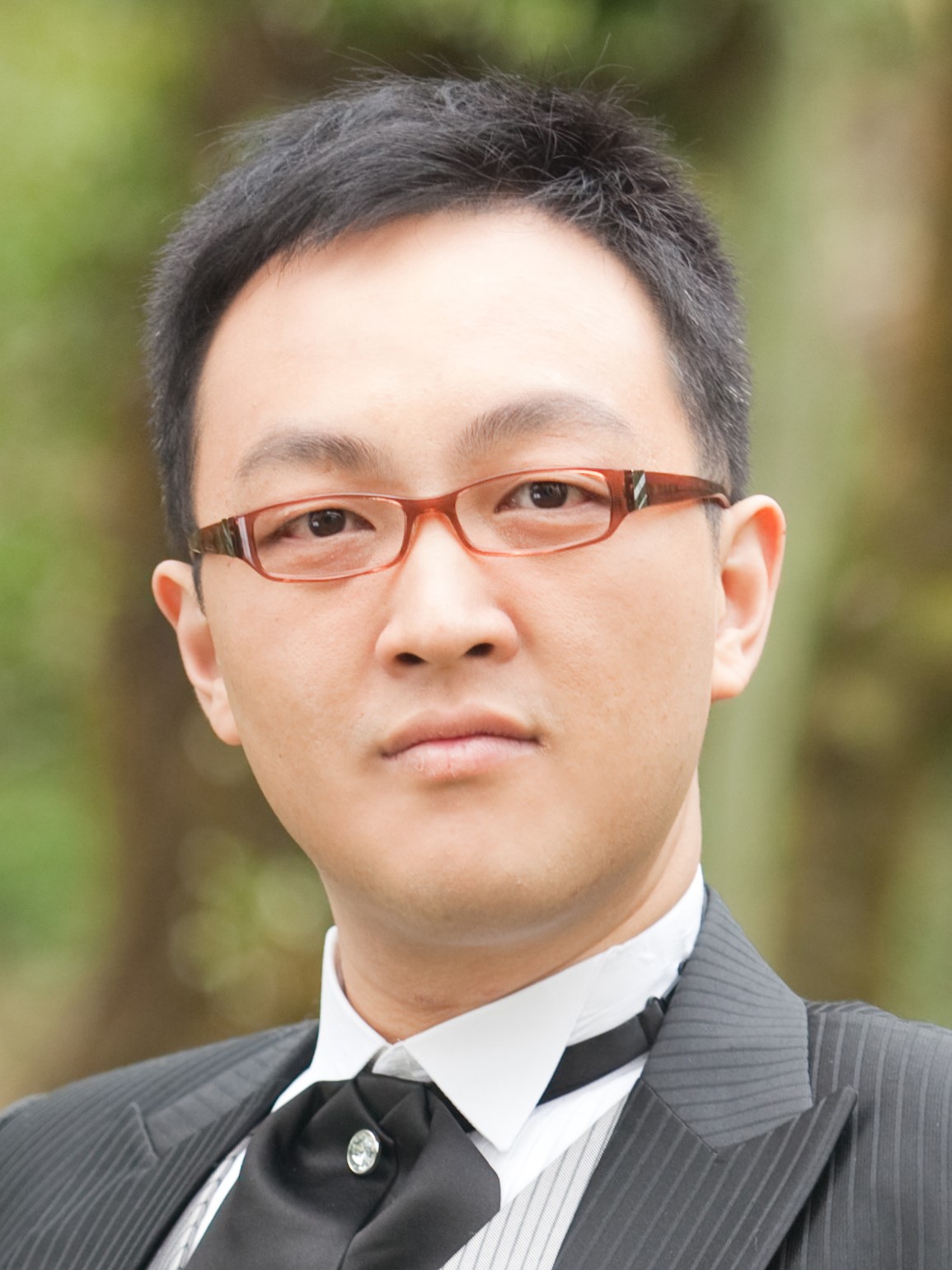}}]{Hwai-Jung Hsu} received his B.E., M.S., and Ph.D. degrees in Computer Science from National Chiao Tung University, Taiwan in 2001, 2003, and 2011. He is now the director of the artificial intelligence reasearch center and an associate professor in the Department of Information Engineering and Computer Science at Feng Chia University. His research interests include data analytics , computer vision and deep learning, agile development,and psychophysiology.
\end{IEEEbiography}

\begin{IEEEbiography}[{\includegraphics[width=1in,height=1.25in,clip,keepaspectratio]{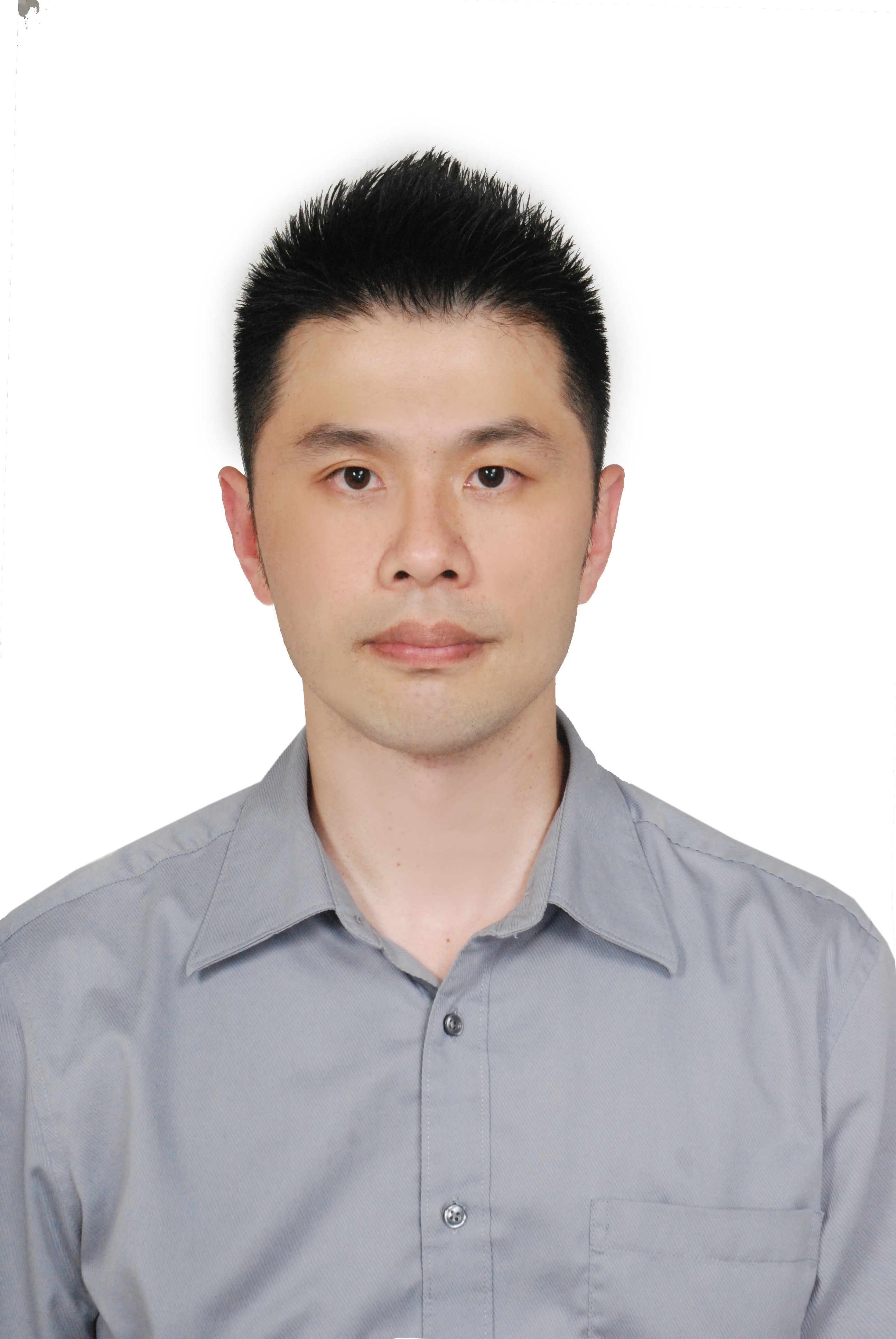}}]{Che-Wei Chou} is an assistant professor at the Department of Industrial Engineering and Systems Management at Feng Chia University. He received a Ph.D. degree in Industrial Engineering and Engineering Management from National Tsing Hua University, Taiwan, in 2016. He also received a B.S. degree and M.S. in physics from Industrial Engineering Management from Yuan Ze University, Taiwan, in 2001 and 2003, respectively. He served as the CEO of Artificial Intelligence for Intelligent Manufacturing Systems (AIMS) Research Center, MOST, Taiwan from 2019 to 2021. His research interests include intelligent manufacturing, manufacturing large language models, deep learning, decision analysis, and big data analytics.
\end{IEEEbiography}

\begin{IEEEbiography}[{\includegraphics[width=1in,height=1.25in,clip,keepaspectratio]{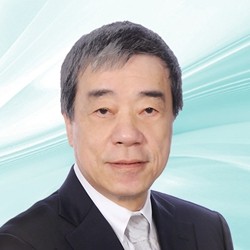}}]{H. T. Kung} is William H. Gates Professor of Computer Science and Electrical Engineering at Harvard University. He conducts research on topics related to the application of artificial intelligence in manufacturing and healthcare, AI accelerators, VLSI design, high-performance computing, parallel and distributed computing, computer architectures, and computer networks.

Professor Kung received his bachelor's degree from National Tsing Hua University in Taiwan in 1968. He received his Ph.D. degree from Carnegie Mellon University in 1973. He taught at Carnegie Mellon for 19 years before joining Harvard in 1992.

Professor Kung is an elected member of the US National Academy of Engineering for introducing the idea of systolic computation, contributions to parallel computing, and applying complexity analysis to very-large-scale-integrated (VLSI) computation. In addition, Professor Kung is an ACM Fellow, Guggenheim Fellow, an elected member of the Academia Sinica in Taiwan, as well as the president and co-founder of the Taiwan AI Academy, a non-profit organization which has cultivated over 10,000 AI talents for industries since 2019.
\end{IEEEbiography}

\begin{IEEEbiography}[{\includegraphics[width=1in,height=1.25in,clip,keepaspectratio]{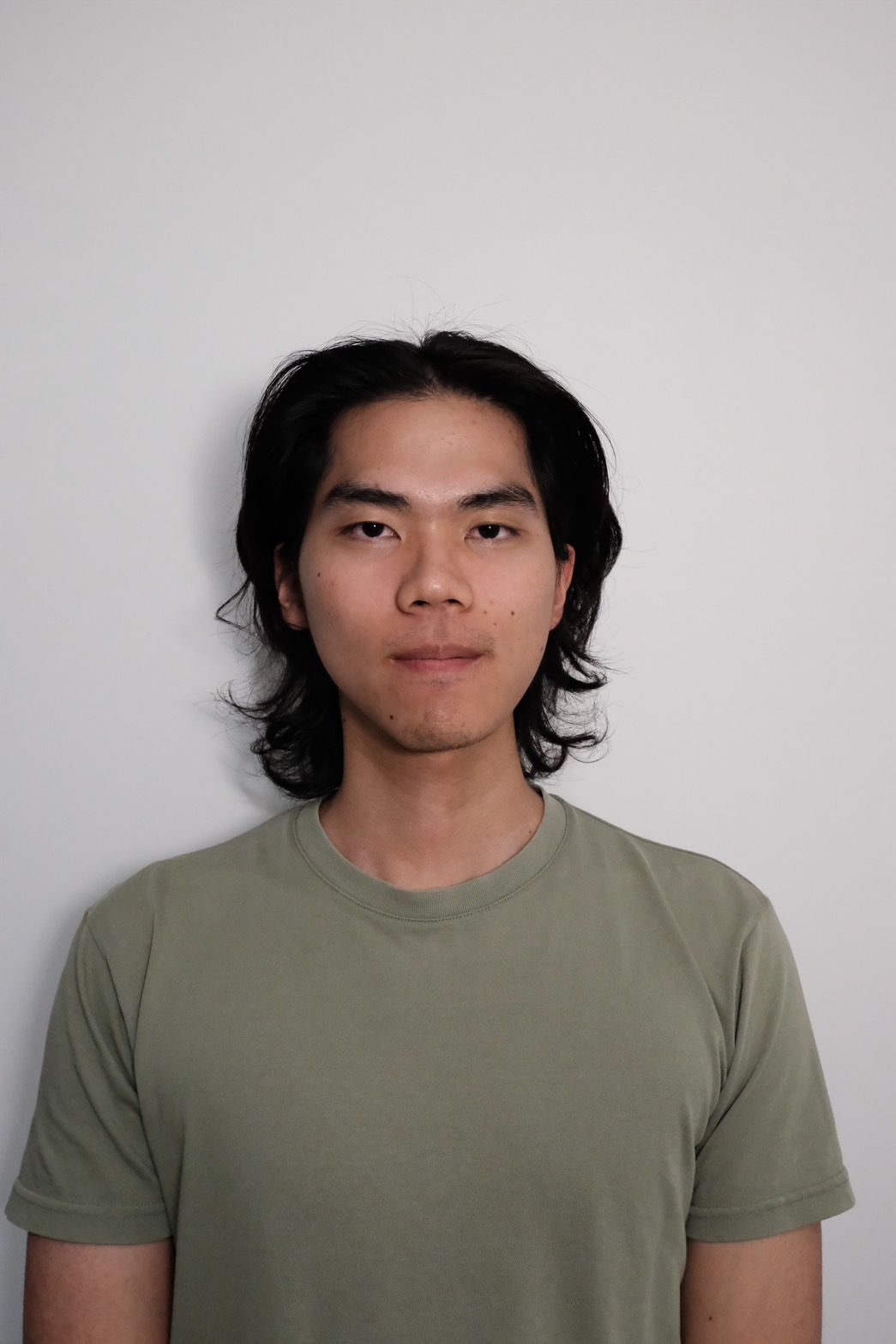}}]{Chen-Hsin Lee} received his Duo-Bachelor’s Degree in Business Management, with a major in Business Information Systems, in 2023 from The University of Queensland, Australia, and Feng Chia University, Taiwan. He is currently pursuing a Master’s Degree in Information Technology, specializing in Data Analytics at The University of Technology Sydney, Australia.
\end{IEEEbiography}

\begin{IEEEbiography}[{\includegraphics[width=1in,height=1.25in,clip,keepaspectratio]{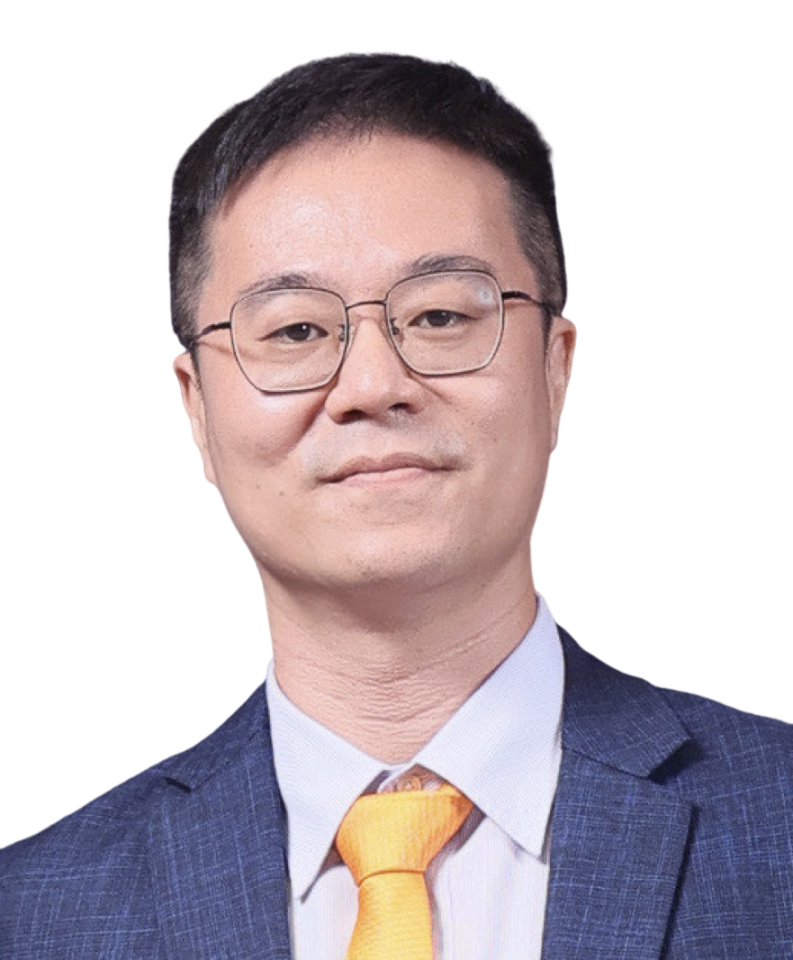}}]{Sheng-Mao Cheng} received two M.S. Degree in Physics and Industrial Engineering
and Engineering Management from National Tsing Hua University, Taiwan, in 2011 and 2022. He worked as a process engineer at TSMC for 3 years from 2012 to 2015. 
From 2015 he switched to the precise machine industry. Currently, he is the manager of Victor Taichung Machinery Company. 
He has participated in many important major projects of Victor Taichung, such as production line improvement, information system integration, and new generation products and application technology development.

\end{IEEEbiography}

\end{document}